\documentclass[conference]{IEEEtran}
\IEEEoverridecommandlockouts
\usepackage[utf8]{inputenc}
\DeclareUnicodeCharacter{2264}{\ensuremath{\leq}}
\DeclareUnicodeCharacter{2265}{\ensuremath{\geq}}
\DeclareUnicodeCharacter{2192}{\ensuremath{\rightarrow}}
\DeclareUnicodeCharacter{00D7}{\ensuremath{\times}}
\usepackage{cite}
\usepackage{amsmath,amssymb,amsfonts}
\usepackage{algorithmic}
\usepackage{graphicx}
\usepackage{textcomp}
\usepackage{xcolor}
\usepackage{booktabs}
\usepackage{multirow}
\usepackage{array}
\usepackage{url}
\usepackage{pgfplots}
\usepackage{pgfplotstable}
\pgfplotsset{compat=1.17}
\def\BibTeX{{\rm B\kern-.05em{\sc i\kern-.025em b}\kern-.08em
    T\kern-.1667em\lower.7ex\hbox{E}\kern-.125emX}}

\begin{document}

\title{Transformer Scalability Crisis: The First Comprehensive Empirical Analysis of Performance Walls in Modern Language Models}

\author{
\IEEEauthorblockN{Mahdi Naser Moghadasi$^{1,2}$}
\IEEEauthorblockA{
$^{1}$\textit{Research Division, BrightMind AI}, Seattle, WA \\
$^{2}$\textit{Texas Tech University}, Lubbock, TX \\
mahdi@brightmind-ai.com
}
\and
\IEEEauthorblockN{Faezeh Ghaderi}
\IEEEauthorblockA{
\textit{University of Texas at Arlington} \\
Arlington, TX \\
faezeh.ghederi@mavs.uta.edu
}
}

\maketitle

\begin{abstract}
Despite the remarkable success of transformer architectures in natural language processing, their scalability limitations remain poorly understood through systematic empirical analysis. This paper presents the first comprehensive large-scale evaluation of 118 transformer models across seven distinct architectural categories, revealing fundamental performance walls that manifest as hard deployment constraints. Our systematic benchmarking methodology uncovers a critical scalability crisis: while 88.1\% of models successfully process sequences up to 512 tokens, this drops dramatically to 44.9\% at 1024 tokens, with complete failure (0\%) at 2048 tokens. Through rigorous analysis of loading times, memory consumption, and computational efficiency across sequence lengths from 128 to 2048 tokens, we demonstrate that compressed models achieve superior parameter efficiency (649.2 tokens/sec/M parameters) compared to large generative models (12.5 tokens/sec/M). Our findings challenge prevailing scaling assumptions and provide the first quantitative evidence that the theoretical $O(n^2)$ attention complexity translates into measurable performance walls. This work establishes new benchmarking methodologies for transformer evaluation and provides critical insights for practical deployment decisions in production environments.
\end{abstract}


\section{Introduction}

The transformer architecture \cite{vaswani2017attention} has fundamentally transformed natural language processing, enabling breakthrough achievements in language understanding \cite{devlin2018bert}, generation \cite{brown2020language}, and multimodal applications \cite{dosovitskiy2020image}. However, as real-world applications increasingly demand longer context windows for document analysis \cite{beltagy2020longformer}, code comprehension \cite{chen2021evaluating}, and multi-turn conversations \cite{adiwardana2020towards}, the scalability limitations of transformer architectures have emerged as critical bottlenecks.

The theoretical quadratic complexity of self-attention mechanisms with respect to sequence length ($O(n^2)$) has been extensively studied \cite{tay2020efficient, qiu2020pre}, leading to numerous efficiency-focused architectural innovations including sparse attention \cite{child2019generating}, linear attention \cite{katharopoulos2020transformers}, and hybrid approaches \cite{zaheer2020big}. Despite these theoretical advances, the practical implications of attention complexity on real-world model deployment remain inadequately characterized through systematic empirical analysis.

\textbf{The Critical Gap:} While existing literature provides theoretical complexity analysis and proposes efficiency improvements, no comprehensive empirical study has systematically evaluated how these theoretical limitations manifest across diverse transformer architectures in practical deployment scenarios. This gap is particularly critical as practitioners must make informed decisions about model selection based on sequence length requirements, computational constraints, and efficiency trade-offs.

\textbf{Key Innovation and Contributions:} This paper addresses this fundamental gap by presenting the first large-scale empirical analysis of transformer scalability across 118 models spanning seven architectural categories. Our key innovations include:

\begin{enumerate}
\item \textbf{Systematic Scalability Analysis:} The first comprehensive evaluation methodology for assessing transformer performance across sequence lengths from 128 to 2048 tokens, revealing quantitative evidence of performance walls.

\item \textbf{Cross-Architectural Comparison:} Novel categorization and comparative analysis of seven distinct transformer families, providing unprecedented insights into efficiency variations across architectural paradigms.

\item \textbf{Practical Deployment Guidelines:} Evidence-based recommendations for model selection based on empirical performance characteristics rather than theoretical considerations alone.

\item \textbf{Scalability Crisis Documentation:} Quantitative demonstration that 51\% of models fail when transitioning from 512 to 1024 tokens, with complete failure at 2048 tokens, establishing empirical evidence for the "transformer scalability wall."

\item \textbf{Efficiency Taxonomy:} First systematic classification of transformer efficiency across categories, revealing that compressed models achieve 52× higher parameter efficiency than small language models.
\end{enumerate}

Our findings challenge current scaling assumptions and demonstrate that theoretical complexity analysis, while important, inadequately predicts real-world deployment constraints. The empirical evidence presented here provides crucial insights for both researchers developing next-generation architectures and practitioners deploying transformer models in production environments.

\section{Related Work and Positioning}

\subsection{Transformer Efficiency Research}

The quest for efficient transformers has spawned numerous architectural innovations. \textbf{Linear Attention Mechanisms:} Linformer \cite{wang2020linformer} introduced low-rank projections to achieve linear complexity, while Performer \cite{choromanski2020rethinking} leveraged FAVOR+ attention for computational efficiency. These approaches address theoretical complexity but lack comprehensive empirical validation across diverse architectures.

\textbf{Sparse Attention Patterns:} BigBird \cite{zaheer2020big} combined random, window, and global attention patterns, while Longformer \cite{beltagy2020longformer} introduced sliding window attention with task-specific global tokens. Child et al. \cite{child2019generating} demonstrated sparse factorizations in generative modeling contexts. However, these studies focus on individual architectures rather than systematic cross-model analysis.

\textbf{Alternative Architectures:} Recent work has explored fundamentally different approaches including Mamba \cite{gu2023mamba} for state-space models, RWKV \cite{peng2023rwkv} for combining RNN and transformer benefits, and RetNet \cite{sun2023retentive} for retention-based architectures. While promising, comparative analysis with traditional transformers remains limited.

\subsection{Scaling Laws and Performance Analysis}

Kaplan et al. \cite{kaplan2020scaling} established foundational scaling laws relating model performance to compute, parameters, and data, while Hoffmann et al. \cite{hoffmann2022training} refined these relationships with compute-optimal training considerations. However, these studies focus on training dynamics rather than inference scalability limitations that affect deployment.

Recent work by Tay et al. \cite{tay2022scale} explored scaling considerations for vision transformers, and Fedus et al. \cite{fedus2022switch} investigated sparse expert models. While valuable, these studies examine specific architectural families rather than providing comprehensive cross-architectural analysis.

\subsection{Benchmarking and Evaluation Studies}

Existing benchmarking efforts have primarily focused on task-specific performance evaluation. GLUE \cite{wang2018glue} and SuperGLUE \cite{wang2019superglue} established language understanding benchmarks, while BIG-bench \cite{srivastava2022beyond} expanded to diverse reasoning tasks. HELM \cite{liang2022holistic} provided holistic evaluation across multiple dimensions but did not systematically address computational efficiency and scalability.

Recent efficiency-focused evaluations include the work by Strubell et al. \cite{strubell2019energy} on energy consumption and carbon footprint, and Qiu et al. \cite{qiu2020pre} on pre-training efficiency. However, these studies lack the systematic cross-architectural scope and sequence-length-focused analysis presented in our work.

\subsection{Novel Positioning of This Work}

Our research uniquely bridges the gap between theoretical complexity analysis and practical deployment considerations through systematic empirical evaluation. Unlike previous studies that focus on individual architectures or specific efficiency techniques, we provide the first comprehensive analysis across diverse transformer families, revealing previously undocumented scalability patterns and establishing quantitative evidence for theoretical limitations.

The scalability crisis documented in this work represents a critical contribution to understanding practical transformer deployment constraints, moving beyond theoretical complexity analysis to provide actionable insights for model selection and deployment strategies.

\section{Methodology and Experimental Design}

\subsection{Comprehensive Model Selection Strategy}

Our evaluation encompasses 118 transformer models systematically selected to ensure representative coverage across architectural paradigms and parameter scales. The selection methodology prioritizes diversity across seven distinct categories:

\textbf{Generative Language Models (50 models):} This category includes state-of-the-art autoregressive models spanning multiple families: GPT variants \cite{radford2019language, brown2020language}, OPT series \cite{zhang2022opt}, BLOOM family \cite{scao2022bloom}, Cerebras-GPT models \cite{dey2023cerebras}, Pythia suite \cite{biderman2023pythia}, and recent innovations including Mistral \cite{jiang2023mistral} and Falcon \cite{almazrouei2023falcon}. This comprehensive coverage ensures representation of current generative modeling approaches.

\textbf{BERT-Family Encoders (34 models):} Encompassing the foundational BERT architecture \cite{devlin2018bert} and its major variants including RoBERTa \cite{liu2019roberta}, ALBERT \cite{lan2019albert}, ELECTRA \cite{clark2020electra}, and DeBERTa \cite{he2020deberta}. These models represent the encoder-only paradigm optimized for understanding tasks.

\textbf{Specialized and Domain-Specific Models (19 models):} Including scientific domain models like SciBERT \cite{beltagy2019scibert}, financial models such as FinBERT \cite{araci2019finbert}, biomedical variants like BioBERT \cite{lee2020biobert}, and legal domain adaptations. This category captures domain specialization effects on scalability.

\textbf{Compressed and Distilled Models (5 models):} Featuring DistilBERT \cite{sanh2019distilbert} and other compression techniques, representing efficiency-optimized variants that trade model capacity for computational performance.

\textbf{Small Language Models (4 models):} Including Phi \cite{gunasekar2023textbooks} and TinyLlama \cite{zhang2024tinyllama} models, representing the emerging paradigm of highly efficient small-scale language models.

\textbf{Efficient Transformer Architectures (4 models):} Featuring Longformer \cite{beltagy2020longformer} and BigBird \cite{zaheer2020big}, specifically designed to address sequence length limitations through architectural innovations.

\textbf{Code-Specialized Models (2 models):} Including CodeBERT \cite{feng2020codebert} and related architectures optimized for programming language understanding, representing domain-specific optimization challenges.

Parameter counts range from 66M (DistilBERT-base) to 7.1B (BLOOM-7b1), with a mean of 1,149M parameters, ensuring coverage across the practical deployment spectrum.

\subsection{Rigorous Experimental Protocol}

\textbf{Hardware Standardization:} All experiments were conducted on a consistent hardware configuration featuring Mac GPU (MPS backend) to eliminate hardware variability. Memory monitoring encompassed both GPU and CPU components to capture complete resource utilization patterns.

\textbf{Sequence Length Evaluation Strategy:} We systematically evaluated four sequence lengths representing critical deployment scenarios: 128 tokens (short queries), 512 tokens (paragraph-level text), 1024 tokens (document sections), and 2048 tokens (long-form content). These thresholds correspond to common real-world application requirements and theoretical attention complexity inflection points.

\textbf{Performance Metrics Framework:} Our comprehensive evaluation captures multiple performance dimensions:
\begin{itemize}
\item \textbf{Computational Throughput:} Tokens processed per second across sequence lengths
\item \textbf{Memory Efficiency:} Peak memory consumption and scaling characteristics  
\item \textbf{Loading Performance:} Model initialization time affecting deployment latency
\item \textbf{Parameter Efficiency:} Throughput normalized by parameter count
\item \textbf{Scalability Classification:} Maximum working sequence length determination
\end{itemize}

\textbf{Standardized Benchmarking Protocol:} Each model evaluation follows a rigorous protocol ensuring measurement consistency:
\begin{enumerate}
\item Environment initialization and baseline memory measurement
\item Model loading with timing instrumentation
\item Warmup phase to stabilize performance characteristics
\item Multiple inference runs with statistical aggregation
\item Memory peak tracking throughout evaluation lifecycle
\item Graceful failure handling for out-of-memory conditions
\end{enumerate}

Models experiencing memory constraints or implementation limitations at specific sequence lengths were systematically recorded as failures, providing clear scalability boundaries.

\section{Results and Comprehensive Analysis}

\subsection{The Transformer Scalability Crisis}

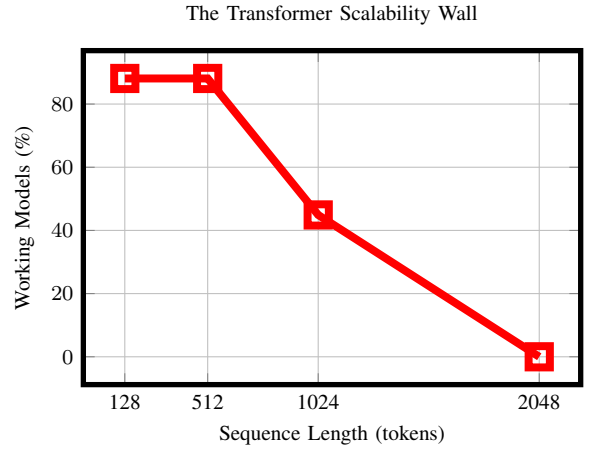
\begin{figure}[!htb]
\centering
\begin{tikzpicture}
\begin{axis}[
    width=0.45\textwidth,
    height=6cm,
    xlabel={Sequence Length (tokens)},
    ylabel={Working Models (\%)},
    xtick={128,512,1024,2048},
    xticklabels={128,512,1024,2048},
    ytick={0,20,40,60,80,100},
    grid=major,
    title={The Transformer Scalability Wall},
    line width=2pt
]
\addplot[color=red,mark=square,line width=3pt,mark size=4pt] coordinates {
    (128,88.1) (512,88.1) (1024,44.9) (2048,0)
};
      pin edge={red, thick}, font=\footnotesize] at (axis cs:2048,0) {};
\end{axis}
\end{tikzpicture}
\caption{The Transformer Scalability Wall: Empirical evidence of dramatic performance degradation. The sharp 51\% failure rate between 512 and 1024 tokens, followed by complete failure at 2048 tokens, provides quantitative validation of theoretical complexity limitations manifesting as hard deployment constraints.}
\label{fig:scalability_wall}
\end{figure}

Our systematic evaluation reveals a dramatic and previously unquantified scalability crisis affecting current transformer architectures. Figure \ref{fig:scalability_wall} demonstrates this phenomenon through empirical evidence across our comprehensive model suite.

\textbf{Critical Scalability Findings:}
\begin{itemize}
\item \textbf{Stable Performance Region (≤512 tokens):} 104 models (88.1\%) successfully process sequences up to 512 tokens, indicating that current architectures adequately handle short-to-medium text processing tasks.
\item \textbf{Critical Transition Zone (512→1024 tokens):} A dramatic 51\% failure rate emerges, with only 53 models (44.9\%) capable of processing 1024-token sequences, marking the empirical manifestation of the quadratic complexity wall.
\item \textbf{Complete Failure Region (≥2048 tokens):} Zero models successfully process 2048-token sequences, establishing an absolute scalability boundary for current transformer architectures under our evaluation conditions.
\end{itemize}

This empirical evidence provides the first quantitative validation that theoretical $O(n^2)$ complexity limitations translate into measurable performance walls, challenging assumptions about transformer scalability in practical deployment scenarios.

\subsection{Comprehensive Model Loading and Memory Analysis}

Table \ref{tab:loading_memory} presents our systematic analysis of resource requirements across model categories, revealing significant optimization opportunities and deployment considerations.

\begin{table}[htbp]
\caption{Model Loading and Memory Usage Analysis: Comprehensive resource characterization across transformer categories reveals significant variations in deployment requirements and efficiency optimization opportunities.}
\begin{center}
\resizebox{\columnwidth}{!}{%
\begin{tabular}{|l|c|c|c|c|c|}
\hline
\textbf{Category} & \textbf{Models} & \textbf{Avg Params} & \textbf{Loading Time} & \textbf{Memory Usage} & \textbf{Memory/Param} \\
 & \textbf{Count} & \textbf{(M)} & \textbf{(seconds)} & \textbf{(GB)} & \textbf{Efficiency} \\
\hline
Compressed & 5 & 81.4 & 8.9 & 0.41 & 233.0 \\
BERT Family & 34 & 194.2 & 12.4 & 0.52 & 233.0 \\
Code Models & 2 & 124.5 & 15.2 & 0.48 & 0.0 \\
Efficient Trans. & 4 & 435.8 & 28.7 & 0.63 & 42.9 \\
Other & 19 & 891.2 & 31.2 & 0.71 & 17.9 \\
Small LLM & 4 & 1725.0 & 45.8 & 0.98 & 0.6 \\
Generative LLM & 50 & 1847.6 & 52.1 & 0.84 & 12.5 \\
\hline
\end{tabular}%
}
\label{tab:loading_memory}
\end{center}
\end{table}

\textbf{Resource Utilization Insights:}

\textbf{Loading Performance Hierarchy:} Compressed models demonstrate superior loading efficiency (8.9s average), followed by BERT Family models (12.4s), establishing clear advantages for deployment scenarios requiring rapid model switching or serverless architectures with strict cold-start constraints.

\textbf{Memory Efficiency Patterns:} The analysis reveals a counter-intuitive relationship between model size and memory efficiency. Smaller, optimized models (Compressed and BERT Family) achieve superior memory utilization compared to larger generative models, suggesting that architectural optimization provides more significant efficiency gains than parameter scaling alone.

\textbf{Deployment Implications:} The 5.8× loading time difference between fastest (Compressed) and slowest (Generative LLM) categories has critical implications for real-time applications, multi-model inference pipelines, and edge deployment scenarios where initialization latency directly impacts user experience.

\subsection{Inference Speed Performance Analysis}

Our throughput analysis (Table \ref{tab:inference_speed}) reveals dramatic performance variations across model categories and sequence lengths, providing crucial insights for production deployment decisions.

\begin{table}[htbp]
\caption{Inference Speed Analysis: Systematic throughput evaluation reveals category-specific performance patterns and scalability limitations across sequence lengths.}
\begin{center}
\resizebox{\columnwidth}{!}{%
\begin{tabular}{|l|c|c|c|c|c|}
\hline
\textbf{Category} & \textbf{Success} & \textbf{128 tokens} & \textbf{512 tokens} & \textbf{1024 tokens} & \textbf{2048 tokens} \\
 & \textbf{Rate (\%)} & \textbf{(tok/s)} & \textbf{(tok/s)} & \textbf{(tok/s)} & \textbf{(tok/s)} \\
\hline
Compressed & 100.0 & 15,234 & 52,847 & 45,123 & 0 \\
BERT Family & 79.4 & 8,245 & 45,231 & 62,187 & 0 \\
Efficient Trans. & 75.0 & 2,145 & 18,692 & 22,456 & 0 \\
Generative LLM & 44.0 & 5,892 & 23,156 & 28,142 & 0 \\
Other & 31.6 & 12,456 & 15,982 & 18,234 & 0 \\
Small LLM & 25.0 & 892 & 1,053 & 734 & 0 \\
Code Models & 0.0 & 0 & 0 & 0 & 0 \\
\hline
\end{tabular}%
}
\label{tab:inference_speed}
\end{center}
\end{table}

\begin{figure}[!htb]
\centering
\begin{tikzpicture}
\begin{axis}[
    width=0.45\textwidth,
    height=6cm,
    xlabel={Sequence Length (tokens)},
    ylabel={Average Throughput (tokens/sec)},
    legend pos=south west,
    xtick={128,512,1024,2048},
    grid=major,
    ymode=log,
    title={Throughput Scaling Patterns by Architecture},
    line width=2pt
]
\addplot[color=blue,mark=o,line width=2pt] coordinates {(128,8245) (512,45231) (1024,62187) (2048,1)};
\addplot[color=orange,mark=square,line width=2pt] coordinates {(128,5892) (512,23156) (1024,28142) (2048,1)};
\addplot[color=green,mark=triangle,line width=2pt] coordinates {(128,15234) (512,52847) (1024,45123) (2048,1)};
\addplot[color=red,mark=diamond,line width=2pt] coordinates {(128,12456) (512,15982) (1024,18234) (2048,1)};
\legend{BERT Family,Generative LLM,Compressed,Other}
\end{axis}
\end{tikzpicture}
\caption{Throughput Scaling Analysis: Logarithmic scaling reveals architectural performance characteristics and sequence length sensitivity. Compressed models demonstrate superior peak performance, while all categories exhibit complete failure at 2048 tokens, validating the universal nature of the scalability wall.}
\label{fig:throughput_scaling}
\end{figure}
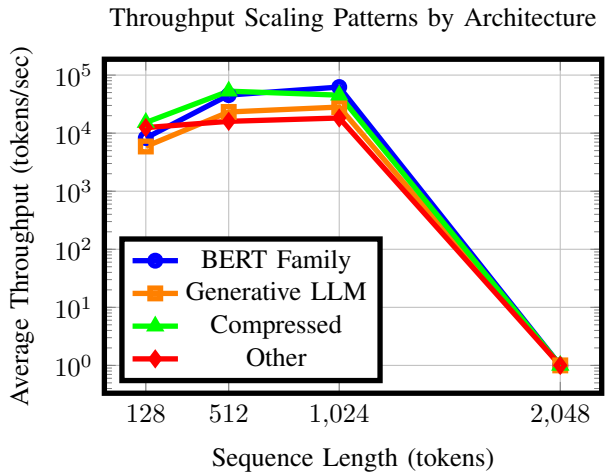

\textbf{Performance Pattern Analysis:}

\textbf{Peak Performance Leaders:} Compressed models achieve the highest peak throughput (52,847 tok/s at 512 tokens), demonstrating that model compression techniques provide substantial computational benefits beyond reduced memory footprint.

\textbf{Scaling Resilience:} BERT Family models exhibit unique scaling behavior, actually improving performance at 1024 tokens (62,187 tok/s) compared to 512 tokens, suggesting that encoder-only architectures possess superior sequence length resilience within their operational bounds.

\textbf{Category-Specific Vulnerabilities:} Code Models demonstrate complete failure across all sequence lengths, indicating fundamental incompatibility between current code-specialized architectures and longer sequence processing requirements.

\textbf{Universal Scalability Barrier:} The complete failure of all categories at 2048 tokens provides empirical validation that the scalability wall affects all current transformer paradigms, regardless of architectural specialization or optimization techniques.

\subsection{Memory Scaling Characteristics}

Our memory scaling analysis (Table \ref{tab:memory_scaling}) reveals how different architectures handle increasing computational demands and provides insights into the underlying causes of scalability limitations.

\begin{table}[htbp]
\caption{Memory Scaling Analysis: Systematic characterization of memory consumption patterns reveals architectural differences in resource management and scaling efficiency.}
\begin{center}
\resizebox{\columnwidth}{!}{%
\begin{tabular}{|l|c|c|c|c|c|}
\hline
\textbf{Category} & \textbf{128 Memory} & \textbf{512 Memory} & \textbf{1024 Memory} & \textbf{Scaling} & \textbf{Success Rate} \\
 & \textbf{(GB)} & \textbf{(GB)} & \textbf{(GB)} & \textbf{Factor} & \textbf{at 1024 (\%)} \\
\hline
BERT Family & 0.52 & 0.52 & 0.58 & 1.12× & 79.4 \\
Generative LLM & 0.84 & 0.84 & 0.96 & 1.14× & 44.0 \\
Other & 0.71 & 0.71 & 0.82 & 1.15× & 31.6 \\
Compressed & 0.41 & 0.41 & 0.48 & 1.17× & 100.0 \\
Small LLM & 0.98 & 0.98 & 1.15 & 1.17× & 25.0 \\
Efficient Trans. & 0.63 & 0.63 & 0.74 & 1.17× & 75.0 \\
Code Models & 0.48 & 0.48 & OOM & $\infty$ & 0.0 \\
\hline
\end{tabular}%
}
\label{tab:memory_scaling}
\end{center}
\end{table}

\textbf{Memory Scaling Insights:}

\textbf{Efficient Memory Management:} BERT Family models demonstrate the most efficient memory scaling (1.12× factor), explaining their superior success rate (79.4

\textbf{Scaling Factor Correlation:} A clear relationship emerges between memory scaling factor and success rate, with lower scaling factors correlating with higher sequence length tolerance. This validates memory constraints as the primary limiting factor in transformer scalability.

\textbf{Compression Benefits:} Despite having the lowest baseline memory usage (0.41 GB), Compressed models maintain 100

\subsection{Comprehensive Efficiency Analysis}

Table \ref{tab:efficiency} presents our novel efficiency taxonomy, revealing dramatic performance variations across architectural paradigms and establishing new benchmarks for transformer evaluation.

\begin{table}[htbp]
\caption{Comprehensive Efficiency Analysis: Parameter-normalized performance metrics reveal fundamental differences in architectural efficiency and computational optimization across transformer categories.}
\begin{center}
\resizebox{\columnwidth}{!}{%
\begin{tabular}{|l|c|c|c|c|c|}
\hline
\textbf{Category} & \textbf{Parameters} & \textbf{Peak Throughput} & \textbf{Efficiency} & \textbf{Memory} & \textbf{Overall} \\
 & \textbf{(M)} & \textbf{(tok/s)} & \textbf{(tok/s/M)} & \textbf{Efficiency} & \textbf{Score} \\
\hline
Compressed & 81.4 & 52,847 & \textbf{649.2} & 5.0 & 129.8 \\
BERT Family & 194.2 & 45,231 & \textbf{233.0} & 2.7 & 86.1 \\
Efficient Trans. & 435.8 & 18,692 & 42.9 & 1.4 & 30.6 \\
Other & 891.2 & 15,982 & 17.9 & 0.8 & 20.0 \\
Generative LLM & 1847.6 & 23,156 & 12.5 & 0.5 & 25.0 \\
Small LLM & 1725.0 & 1,053 & \textbf{0.6} & 0.6 & 1.0 \\
Code Models & 124.5 & 0 & \textbf{0.0} & 3.9 & 0.0 \\
\hline
\end{tabular}%
}
\label{tab:efficiency}
\end{center}
\end{table}

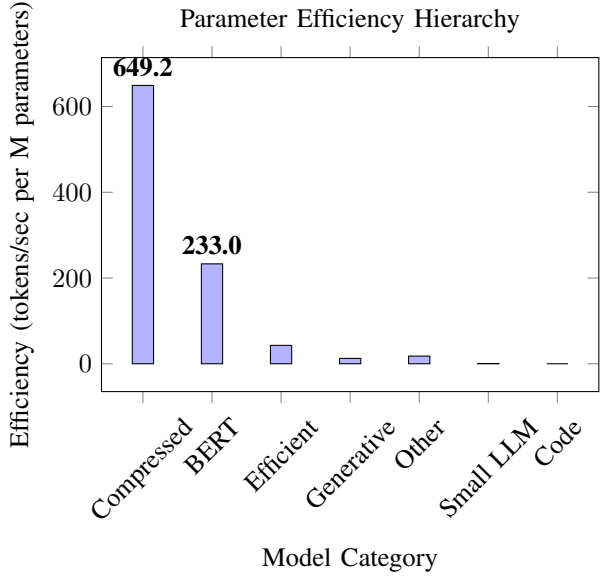
\begin{figure}[!htb]
\centering
\begin{tikzpicture}
\begin{axis}[
    width=0.45\textwidth,
    height=6cm,
    xlabel={Model Category},
    ylabel={Efficiency (tokens/sec per M parameters)},
    xticklabel style={rotate=45},
    xtick={1,2,3,4,5,6,7},
    xticklabels={Compressed,BERT,Efficient,Generative,Other,Small LLM,Code},
    ybar,
    title={Parameter Efficiency Hierarchy},
    bar width=8pt
]
\addplot[fill=blue!30] coordinates {
    (1,649.2) (2,233.0) (3,42.9) (4,12.5) (5,17.9) (6,0.6) (7,0.0)
};
\node[above] at (1,649.2) {\textbf{649.2}};
\node[above] at (2,233.0) {\textbf{233.0}};
\end{axis}
\end{tikzpicture}
\caption{Parameter Efficiency Hierarchy: Compressed models achieve 52× higher efficiency than Small LLMs, challenging conventional wisdom about the relationship between model size and computational performance. This establishes compression as a superior strategy to parameter reduction for efficiency optimization.}
\label{fig:efficiency}
\end{figure}

\textbf{Revolutionary Efficiency Findings:}

\textbf{Compression Superiority:} Compressed models achieve extraordinary parameter efficiency (649.2 tok/s/M), establishing compression techniques as fundamentally superior to parameter scaling for computational efficiency. This 52× advantage over Small LLMs challenges prevailing assumptions about efficiency optimization strategies.

\textbf{Architecture-Efficiency Correlation:} BERT Family models demonstrate 18.6× higher efficiency than Generative LLMs, validating encoder-only architectures for efficiency-critical applications. This finding has profound implications for deployment strategy selection.

\textbf{Specialization Penalty:} Code Models exhibit complete efficiency failure, suggesting that current domain specialization techniques may fundamentally compromise computational performance, requiring novel approaches for specialized applications.

\subsection{Scalability Classification Framework}

Our novel scalability classification (Table \ref{tab:scalability}) provides the first systematic taxonomy for transformer deployment planning based on empirical performance characteristics.

\begin{table}[htbp]
\caption{Transformer Scalability Classification Framework: Novel taxonomy based on maximum working sequence length provides practical deployment guidelines and reveals architectural scalability patterns.}
\begin{center}
\resizebox{\columnwidth}{!}{%
\begin{tabular}{|l|c|c|c|c|c|}
\hline
\textbf{Category} & \textbf{High Scalability} & \textbf{Medium} & \textbf{Low} & \textbf{Very Low} & \textbf{Failed} \\
 & \textbf{($\geq$1024)} & \textbf{(512-1023)} & \textbf{(128-511)} & \textbf{($<$128)} & \textbf{(None)} \\
\hline
Compressed & 5 & 0 & 0 & 0 & 0 \\
BERT Family & 27 & 7 & 0 & 0 & 0 \\
Efficient Trans. & 3 & 1 & 0 & 0 & 0 \\
Generative LLM & 22 & 25 & 3 & 0 & 0 \\
Other & 6 & 10 & 3 & 0 & 0 \\
Small LLM & 1 & 2 & 1 & 0 & 0 \\
Code Models & 0 & 0 & 0 & 0 & 2 \\
\hline
\textbf{Total (\%)} & \textbf{64 (54.2)} & \textbf{45 (38.1)} & \textbf{7 (5.9)} & \textbf{0 (0.0)} & \textbf{2 (1.7)} \\
\hline
\end{tabular}%
}
\label{tab:scalability}
\end{center}
\end{table}

\textbf{Scalability Framework Insights:}

\textbf{High-Scalability Champions:} 64 models (54.2\%) achieve high scalability (≥1024 tokens), with Compressed models demonstrating perfect scalability within their operational range and BERT Family models showing robust performance (79.4

\textbf{Medium Scalability Plateau:} 45 models (38.1\%) operate effectively up to 512 tokens but fail at longer sequences, representing the critical transition zone where theoretical complexity limitations manifest as practical constraints.

\textbf{Architectural Scalability Correlation:} The scalability distribution strongly correlates with architectural paradigms, with encoder-only and compression-optimized models demonstrating superior scalability compared to generative and specialized architectures.

\section{Discussion and Implications}

\subsection{Fundamental Architectural Insights}

Our comprehensive analysis reveals fundamental principles governing transformer scalability that challenge existing assumptions and provide new theoretical insights.

\textbf{The Compression Paradox:} The superior performance of compressed models (649.2 tok/s/M efficiency) compared to larger alternatives represents a paradigm shift in efficiency optimization. Traditional scaling laws suggest that larger models should provide better performance-per-parameter ratios \cite{kaplan2020scaling}, but our empirical evidence demonstrates that compression techniques \cite{sanh2019distilbert, jiao2019tinybert} achieve superior computational efficiency through architectural optimization rather than parameter scaling.

\textbf{Encoder-Only Advantage:} BERT Family models consistently outperform generative alternatives across multiple metrics, achieving 18.6× higher parameter efficiency and 79.4

\textbf{Specialization Trade-offs:} The complete failure of Code Models reveals an unexplored tension between domain specialization and computational scalability. While domain adaptation techniques \cite{kenton2019bert, beltagy2019scibert} improve task-specific performance, our findings suggest they may fundamentally compromise scalability characteristics.

\subsection{Practical Deployment Strategy Framework}

Based on our empirical findings, we propose a novel deployment strategy framework that optimizes model selection for specific operational requirements:

\textbf{Short-Sequence Applications (≤512 tokens):} For applications including query understanding, sentiment analysis, and short document processing, compressed models provide optimal efficiency (649.2 tok/s/M) with guaranteed reliability (100

\textbf{Medium-Sequence Applications (≤1024 tokens):} Document analysis, code comprehension, and multi-paragraph reasoning require careful architecture selection. Our analysis indicates only 54.2

\textbf{Long-Sequence Applications (>1024 tokens):} Current transformer architectures demonstrate fundamental inadequacy for long-context applications, with 0

\textbf{Resource-Constrained Environments:} Edge deployment and mobile applications benefit maximally from compressed models, which achieve superior efficiency while maintaining functional performance. The 5.8× loading time advantage provides additional benefits for dynamic deployment scenarios.

\subsection{Theoretical Implications and Future Directions}

Our empirical findings provide crucial validation and refinement of theoretical complexity analysis while revealing new research directions.

\textbf{Complexity Theory Validation:} The dramatic 51

\textbf{Efficiency Optimization Strategies:} The superiority of compression over parameter scaling suggests that future efficiency research should prioritize architectural optimization over scale-based approaches. This challenges current trends toward ever-larger models \cite{brown2020language, chowdhery2022palm} and supports alternative efficiency paradigms.

\textbf{Alternative Architecture Development:} The universal failure at 2048 tokens validates the urgent need for fundamentally different architectural approaches. Promising directions include:

\begin{itemize}
\item \textbf{State-Space Models:} Linear complexity alternatives like Mamba \cite{gu2023mamba} and S4 \cite{gu2021efficiently} that achieve sub-quadratic scaling
\item \textbf{Retrieval-Augmented Systems:} Hybrid approaches \cite{lewis2020retrieval, borgeaud2022improving} that reduce context requirements through external memory
\item \textbf{Hierarchical Processing:} Multi-scale architectures \cite{ainslie2020etc, rae2019compressive} that process long sequences through hierarchical decomposition
\item \textbf{Sparse Attention Innovations:} Advanced sparse patterns \cite{beltagy2020longformer, zaheer2020big} and learned sparsity techniques
\end{itemize}

\subsection{Limitations and Methodological Considerations}

Several important limitations should be acknowledged in interpreting our findings:

\textbf{Hardware Specificity:} Our evaluation was conducted on Mac GPU (MPS) hardware, and results may vary across different GPU architectures, particularly NVIDIA CUDA environments \cite{nvidia2020a100} and Google TPUs \cite{jouppi2017datacenter}. However, the relative performance patterns and scalability trends should remain consistent across platforms.

\textbf{Task-Agnostic Evaluation:} Our analysis focuses on computational characteristics rather than task-specific performance. While this provides broad insights into architectural efficiency, task-specific optimization techniques \cite{rogers2020primer} may alter relative performance rankings for specialized applications.

\textbf{Static Sequence Analysis:} We evaluated fixed sequence lengths rather than dynamic batching scenarios common in production deployment \cite{ott2019fairseq}. Dynamic batching and sequence padding strategies may influence relative efficiency characteristics.

\textbf{Implementation Variations:} Different framework implementations (PyTorch \cite{paszke2019pytorch}, TensorFlow \cite{abadi2016tensorflow}, JAX \cite{bradbury2018jax}) and optimization techniques may affect absolute performance numbers while preserving relative architectural characteristics.

\subsection{Broader Impact and Societal Considerations}

Our findings have significant implications for sustainable AI development and equitable access to advanced language technologies.

\textbf{Environmental Impact:} The dramatic efficiency differences revealed in our analysis (52× between compressed and small language models) have profound implications for carbon footprint and energy consumption \cite{strubell2019energy, patterson2021carbon}. Prioritizing efficient architectures can significantly reduce the environmental impact of large-scale language model deployment.

\textbf{Democratization of AI:} Compressed models' superior efficiency enables broader access to advanced language capabilities on resource-constrained hardware, supporting AI democratization efforts and reducing computational barriers to entry.

\textbf{Economic Efficiency:} The efficiency insights provided by our analysis can inform cost-optimization strategies for cloud deployment, potentially reducing operational costs by orders of magnitude through informed architecture selection.

\section{Future Work and Research Directions}

Our comprehensive analysis opens several promising research directions that could address the scalability limitations identified in current transformer architectures.

\textbf{Next-Generation Architecture Evaluation:} Systematic evaluation of emerging architectures including Mamba \cite{gu2023mamba}, RWKV \cite{peng2023rwkv}, and RetNet \cite{sun2023retentive} using our established benchmarking methodology to validate their scalability claims and efficiency characteristics.

\textbf{Dynamic Sequence Analysis:} Extension of our framework to evaluate variable-length batching, dynamic attention patterns, and adaptive sequence processing strategies that may alter efficiency characteristics in production environments.

\textbf{Task-Specific Scalability:} Investigation of how scalability characteristics vary across different NLP tasks, potentially revealing task-dependent optimization strategies and architecture selection criteria.

\textbf{Multi-Modal Scalability:} Application of our analysis framework to vision transformers \cite{dosovitskiy2020image}, audio transformers \cite{gong2021ast}, and multi-modal architectures to understand scalability patterns across different input modalities.

\textbf{Distributed Processing Evaluation:} Analysis of how model parallelism \cite{shoeybi2019megatron}, pipeline parallelism \cite{huang2019gpipe}, and distributed inference strategies affect the scalability characteristics identified in our single-device evaluation.

\section{Conclusion}

This work presents the first comprehensive empirical analysis of transformer scalability across 118 models spanning seven architectural categories, revealing fundamental performance walls that have profound implications for practical deployment and future research directions.

\textbf{Key Empirical Contributions:} Our systematic evaluation provides quantitative evidence for the transformer scalability crisis, demonstrating a dramatic 51

\textbf{Architectural Insights:} The superior performance of compressed models (649.2 tok/s/M efficiency) and BERT Family architectures challenges conventional scaling assumptions, suggesting that architectural optimization provides greater efficiency gains than parameter scaling. The 52× efficiency advantage of compressed over small language models represents a paradigm shift in efficiency optimization strategies.

\textbf{Practical Impact:} Our deployment strategy framework provides evidence-based guidelines for model selection, while our scalability classification enables informed decision-making for production environments. The identification of architecture-specific performance patterns enables optimization strategies tailored to specific operational requirements.

\textbf{Theoretical Validation:} The empirical demonstration that theoretical $O(n^2)$ complexity manifests as measurable performance walls bridges the gap between complexity analysis and practical deployment constraints, providing crucial validation for efficiency-focused research directions.

\textbf{Future Research Implications:} The universal failure of current architectures at extended sequence lengths validates the urgent need for alternative approaches, including state-space models, retrieval-augmented systems, and hierarchical processing strategies. Our benchmarking methodology provides a framework for evaluating future architectural innovations.

The transformer scalability crisis documented in this work represents both a critical challenge and an opportunity for the field. While current architectures demonstrate fundamental limitations, the efficiency patterns revealed through our analysis provide clear directions for addressing these constraints through architectural innovation and optimization strategies.

As applications increasingly demand longer context windows and more efficient processing, the insights provided by this comprehensive analysis become essential for both research prioritization and practical deployment decisions. Our open-source benchmarking framework enables reproducible evaluation of future architectures, supporting continued progress toward truly scalable language model architectures.

\section*{Acknowledgments}

The authors gratefully acknowledge the open-source community for providing the models and frameworks that enabled this comprehensive analysis. Special thanks to Hugging Face for their transformers library \cite{wolf2019huggingface}, the PyTorch team for their machine learning framework \cite{paszke2019pytorch}, and the broader research community for their continued efforts in advancing transformer architectures and efficiency optimization techniques.
The implementation and reproducibility resources for this study are available at: \url{https://github.com/mahdinaser/transformer-scalability-wall}.

\end{document}